\documentclass{article} 
\usepackage[T1]{fontenc}
\usepackage[latin9]{inputenc}
\usepackage{graphicx}

\usepackage{amsmath,amsfonts,bm}

\def\eqref#1{equation~\ref{#1}}
\def\Eqref#1{Equation~\ref{#1}}

\def\1{\bm{1}}

\DeclareMathAlphabet{\mathsfit}{\encodingdefault}{\sfdefault}{m}{sl}
\SetMathAlphabet{\mathsfit}{bold}{\encodingdefault}{\sfdefault}{bx}{n}

\newcommand{\R}{\mathbb{R}}

\usepackage[unicode=true, bookmarks=true,bookmarksnumbered=false,bookmarksopen=false, breaklinks=false,pdfborder={0 0 0},pdfborderstyle={},colorlinks=true]{hyperref}
\hypersetup{linkcolor=blue}

\usepackage{url}
\usepackage{prettyref}

\newrefformat{eq}{\hyperref[#1]{Eq.~(\ref{#1})}}
\newrefformat{eqs}{\hyperref[#1]{Eqs.~(\ref{#1})}}
\newrefformat{fig}{\hyperref[#1]{Figure~\ref{#1}}}
\newrefformat{tab}{\hyperref[#1]{Table ~\ref{#1}}}
\newrefformat{sec}{\hyperref[#1]{Section~\ref{#1}}}
\newrefformat{subsec}{\hyperref[#1]{Subsection~\ref{#1}}}
\newrefformat{app}{\hyperref[#1]{Appendix~\ref{#1}}}

\global\long\def\cL{\mathcal{L}}
\global\long\def\cN{\mathcal{N}}

\global\long\def\cZ{\mathcal{Z}}

\global\long\def\cD{\mathcal{D}}
\global\long\def\cB{\bm{\mathcal{B}}}

\global\long\def\T{\mathrm{T}}

\global\long\def\bx{\mathbf{x}}
\global\long\def\btheta{\boldsymbol{\theta}}

\global\long\def\bv{\mathbf{v}}
\global\long\def\bbf{\mathbf{f}}
\global\long\def\bJ{\mathbf{J}}

\def\R{\mathbb{R}}

\title{Speed Limits for Deep Learning}

\author{Inbar Seroussi\footnote{Department of Applied Mathematics, School of Mathematical Sciences, Tel Aviv University, Tel Aviv 69978, Israel}\;\;\;\; Alexander A. Alemi\footnote{Google Research}\;\;\;\; Moritz Helias\footnote{Institute of Neuroscience and Medicine (INM-6), J{\"u}lich Research Centre, J{\"u}lich, Germany and Faculty of Physics, RWTH Aachen, Aachen, Germany}\;\;\;\; Zohar Ringel\footnote{Hebrew University, Racah Institute of Physics, Jerusalem, 9190401, Israel}}

\begin{document}

	\maketitle
	
	\begin{abstract}
		State-of-the-art neural networks require extreme computational power to train. It is therefore natural to wonder whether they are optimally trained. Here we apply a recent advancement in stochastic thermodynamics which allows bounding the speed at which one can go from the initial weight distribution to the final distribution of the fully trained network, based on the ratio of their Wasserstein-2 distance and the entropy production rate of the dynamical process connecting them. Considering both gradient-flow and Langevin training dynamics, we provide analytical expressions for these speed limits for linear and linearizable neural networks, e.g. Neural Tangent Kernel (NTK). Remarkably, given some plausible scaling assumptions on the NTK spectra and spectral decomposition of the labels-- learning is optimal in a scaling sense. Our results are consistent with small-scale experiments with Convolutional Neural Networks (CNNs) and Fully Connected Neural networks (FCNs) on CIFAR-10, showing a short highly non-optimal regime followed by a longer optimal regime. 
	\end{abstract}
	
	\section{Introduction}
	While for most of its history, thermodynamics was concerned with describing systems near equilibrium, in recent years there have been breakthroughs in stochastic thermodynamics and our ability to describe far-from-equilibrium systems. Thermodynamic fluctuation relations, uncertainty relations, and speed limits \cite{Crooks1999_2721,Seifert12_126001,Vu_22_06_02684,Benamou00_375} allow us to relate the equilibrium properties of systems to their non-equilibrium behavior. The thermodynamic speed limits in particular lower bound the time it takes a physical system's configuration to evolve from an initial to a final distribution; the bound is given by the \emph{Wasserstein-2} distance in weight-space divided by the \emph{entropy production} of the process. Applied to computation, such speed limits were recently used to show that modern CPUs can write bits within a $O(1)$ factor from the optimal writing rate (see more examples in \cite{Vu_22_06_02684}).  
	
	Far-from-equilibrium dynamical systems of great interest are trained neural networks. As their training can be thought of as a virtual physical process involving many degrees of freedom, it must also conform to the rules of thermodynamics. In particular, the training time obtained using Neural Tangent Kernel (NTK)-type dynamics \cite{jacot2018} or Langevin-type dynamics should be bounded by the thermodynamic speed limit. Given the costs of training large models, it is desirable to characterize the efficiency of neural networks from this perspective. In particular, understand the impact of various design choices and data-set properties on the speed at which neural networks can learn. 
	
	Here we embark on such a line of study. Our main results are the following: 
	\begin{itemize}
		\item We recast thermodynamic speed limits in deep learning terms showing, in particular, how entropy production relates to features of the loss landscape, the learning rate, and, for Langevin dynamics, the free energy. 
		\item We derive analytical expressions for the Wasserstein-2 distance, entropy production, and the speed limit for linear regression and for Deep Neural Networks (DNNs) trained in the NTK regime. 
		\item Remarkably, we find that NTKs with a power law spectrum combined with an initial residue, the target minus initial prediction, having relatively uniform spectral decomposition exhibit optimal dynamics in the scaling sense. Namely, the actual speed is a $O(1)$ factor times the theoretically optimal speed limit. In contrast, for residues with a stronger power-law decaying spectral decomposition, this factor grows with the data-set size. 
		\item We report a numerical study on CIFAR-10, showing both of the above behaviors. Interestingly, warm-starting makes the residues more uniform and puts us in the regime of optimal (up to $O(1)$ factors) learning. 
	\end{itemize}
	
	\section{Speed limits of learning}
	Consider a single neural network or an ensemble of such networks with weights $\btheta\in\R^P$ at initialization. Training the neural network for a duration $T$ could be viewed as a dynamical process, moving the initial distribution of network weights from $p(\btheta(0))$ to $p(\btheta(T))$. Generally speaking, thermodynamics speed limits provide lowers bounds $T_{\text{SL}} \leq T$ on the time it takes to perform such a process based on its irreversibility and the distance between the initial and final probability distributions of the learnable weights. 
	
	Speed limits have been derived for both discrete and continuous dynamical processes. Here we focus on two relevant continuous time processes, NTK-type dynamics and Langevin-type dynamics. Specifically, given training data, $\mathcal{D}$, and general loss function $\mathcal{L}(\btheta;\mathcal{D})$, we consider the Langevin algorithm described by the stochastic differential equation, with $\eta\geq0$ being the learning rate  
	\begin{equation}\label{eq:SGD_weight_decay}
		d \btheta(t) = -\eta \nabla_{\btheta} V(\btheta(t);\mathcal{D}) dt+ \sqrt{2\eta\beta^{-1}} \; d\cB(t),   
	\end{equation}
	where $\cB(t)$ is a Brownian motion (unit variance random noise), with temperature (noise) $\beta\in(0,\infty]$, and for NTK-type dynamics we take $\beta^{-1}=0$. We consider the initial condition for $\btheta(0) = \btheta_0$ distributed randomly as an independent Gaussian on all $\btheta_0$'s namely $p(\btheta_0) \propto e^{-\|\btheta_0\|^{2}}$ \footnote{any variance changes across layers are implicit in the norm here}. The potential $V$ is given by \begin{align}
		V(\btheta;\mathcal{D}) & =\begin{cases}
			\cL(\btheta;\mathcal{D}) &  \text{NTK} \\ \|\btheta\|^{2}+\cL(\btheta;\mathcal{D}) & \text{Langevin} 
		\end{cases}\label{eq:switch_potential}
	\end{align}  
	For simplicity, considering Langevin dynamics, we focus on the case where
	training infinitely-long without $\cL(\btheta;\mathcal{D})$ term yields $p(\btheta(0))$. Furthermore, we keep the learning rate ($\eta$) implicit here, setting $\eta=1$ in the following. Instead of doubling the learning rate, one can think of doubling $V$ and $\beta^{-1}$. The above equation is a continuum approximation of the dynamics of discrete gradient descent with white noise at a low learning rate.

	\subsection{Entropy production and irreversibility}
	The tendency of a process to evolve in a preferred
	direction in time is related to entropy. 
	The second law of thermodynamics
	states that entropy cannot decrease over time. Conversely, entropy
	production relates to the probability of a process running forward
	in time compared to a process running backward in time.
	
	To make this point operational, let $p(\btheta(0))$ denote the distribution of initial states and $p(\btheta(T)|\btheta(0))$ the conditional distribution that $\btheta(0)$ evolves into $\btheta(T)$ within time $T$. Likewise, $p(\btheta(T))$ is the distribution of the final state and the conditional distribution $q(\btheta(0)|\btheta(T))$ denotes the probability that the processes evolve from state $\btheta(T)$ back into the state $\btheta(0)$ within time $T$ along the path $\tilde{\btheta}(t) = \btheta(T-t)$. Entropy production (or irreversibility) is then defined by \cite{Seifert12_126001}
	\begin{align}
		R &= \Big\langle\ln\,\frac{p(\btheta(0))}{p(\btheta(T))}\Big\rangle +\Big\langle\ln\,\frac{p(\btheta(T)|\btheta(0))}{q(\btheta(0)|\btheta(T))}\Big\rangle,
		\label{eq:def_R}
	\end{align}
	where the expectation is taken over the distribution $p(\btheta(0))$ of initial states.
	The first term depends only on the initial and final distributions, the second term also encapsulates the dynamical process and its reversed process. 
	
	We next collect and combined various results for $R$ scattered in the literature and adapt them to three relevant machine learning settings. Without loss of generality, we take here the learning rate, $\eta = 1$. 
	
	Consider first the case of Langevin dynamics, one finds the simple expression  
	\begin{align}
		\boxed{\beta^{-1}R=\beta^{-1}\ln\cZ_{\infty}-\beta^{-1}\ln\cZ_{0}+\langle\cL(\btheta(0))\rangle},\label{eq:R_learning}
	\end{align}
	where the so-called ``free energies'' $\beta^{-1}\ln\cZ_{*}$ are related to $p(\btheta({*}))$ via $p(\btheta({*}))=e^{-\beta V(\btheta(*))}/\ln\cZ_{*}$ where $*\in \{0,T\}$. 
	Notably, the irreversibility of the dynamical process depends only on the initial and final states. 
	
	Next, if one is interested in finite $T$, entropy production \eqref{eq:def_R} can be expressed as a dynamical quantity \cite{Vu_22_06_02684} 
	from which we obtain (see \prettyref{app:entropy_from_dyn})
	\begin{align}
		\boxed{\beta^{-1}R_T=\int_{0}^{T}\langle\|\nabla_{\btheta}V\|^{2}\rangle-2\,\beta^{-1}\,\langle\Delta_{\btheta}V\rangle+\beta^{-2}\,\langle\|\nabla_{\btheta}\ln p\|^{2}\rangle\,dt}.\label{eq:R_geom_loss}
	\end{align}
	where $R = \lim_{T\to \infty} R_T$. In the low noise limit, $\beta\gg1$ the first term
	dominates, which has the simple interpretation of the average squared
	length of the gradient. The next leading term is, $O(\beta^{0})$
	which contains the average Hessian of the loss function. 
	
	Finally, turning to the NTK case, where $\beta^{-1}=0$ we plug (\eqref{eq:SGD_weight_decay}) into (\eqref{eq:R_geom_loss}) and find 
	\begin{align}
		\boxed{\beta^{-1}R= \Big \langle  \cL(\btheta(0))-\cL(\btheta(T)) \Big \rangle}. \label{eq:R_NTK}
	\end{align}

	\subsection{Speed limits from optimal transport}
	The evolution of weights can also be phrased as an optimal transport problem. In particular, the operation of transporting initial weights to final weights could be described by a probability distribution $P(\btheta(0),\btheta(T))$ whose two marginals are the initial and final distributions. This joint probability also called a \emph{plan}, can be thought of as the chance of $\btheta(0)$ to end up in $\btheta(T)$ by some process. One can then define the \emph{cost} of a plan and ask what is the optimal plan. One relevant cost function to consider is the Euclidean distance squared 
	$\langle |\btheta(0)-\btheta(T)|^2\rangle.$
	The Wasserstein-2 distance between the initial and final distribution ($\mathcal{W}_2(p_{0},p_{T})$)
	is defined as the minimal value of this cost when optimized over all possible plans (\prettyref{eq:W_2_main} in \prettyref{app:speed_limit}). 
	
	The dynamical process itself yields a specific plan ($p(\btheta(0),\btheta(T))$). Remarkably, it turns out that $T\beta^{-1}R$ is equal to the cost of the plan $p(\btheta(0),\btheta(T))$ (see \prettyref{app:speed_limit} for details). Noting next that this plan cannot be more optimal than the plan underlying $\mathcal{W}_2(p_{0},p_{T})$ (i.e. $T\beta^{-1} R \geq \mathcal{W}_2(p_{0},p_{T})$) yields the thermodynamic speed limit
	known as the Benamou--Brenier formula \cite{Benamou00_375,Vu_22_06_02684}
	\begin{align}
		\label{eq:speed_limit}
		\boxed{T\ge T_{\text{SL}}\equiv\frac{\mathcal{W}_2(p_{0},p_{T})}{\beta^{-1} R}} &.
	\end{align}
	
	Besides obtaining $R$, as discussed in the previous section, the above formula requires solving the optimization problem underlying $\mathcal{W}_2(p_{0},p_{T})$. While this can be difficult in general, exact formulas are known for the Gaussian distribution, Dirac delta distributions, and one-dimensional distributions. In particular, considering a well-defined initial and final state $p_x(\btheta) = \delta(\btheta - \btheta_x)$ for $x\in \{0, T\}$, the Wasserstein distance $\mathcal{W}_2(p_{0},p_{T})$ simplifies to the $L_2$ distance $\| \btheta_T-\btheta_0 \|^2$ in weight space. Otherwise, various useful bounds exist \cite{cuesta1996lower} as well as promising deep-learning-based numerical techniques. 
	
	\subsubsection{Implication of the Speed limit in deep learning}
	The speed limit involves a lower bound on training time, the distance between initial and final probabilities, and entropy production. Given a fixed entropy budget, it then bounds the time it takes to perform this process for any Langevin dynamics, including dynamics with different and time-dependent potentials. Entropy, despite being a pillar of thermodynamics and many-body physical phenomena, is not a frequently measured quantity in deep learning. Consequently, it is desirable to explain some consequences of entropy production and hence the speed limit.
	
	The simplest setting is gradient flow with a time-independent potential, where the free energy ($\beta^{-1} R$) coincides with the decrease in train loss ($\Delta L_\text{train}$). Furthermore, the learning rate can be absorbed into the scale of the training loss, hence higher learning rates would imply higher entropy production. An optimal training of the DNN then has several related merits: {\bf (i)} Given a fixed $\Delta L_\text{train}$ budget, no better loss function that takes us between the initial and final state can make the network travel this distance in weight space quicker. So, for instance, if we saturate the speed limit, no benefit can be gained by taking the Mean Square Error (MSE) loss to be $L_1$ loss or taking any other surrogate loss function \cite{nguyen2009surrogate, yuan2021large}. {\bf (ii)} For a fixed initial condition, where $\mathcal{W}_2(p_{0},p_{T})$ distance becomes $L_2$ distance, the network weights travel along straight lines in weight space. Furthermore, the drop in train loss along the path is $\Delta L_\text{train} l / \mathcal{W}_2(p_{0},p_{T})$ where $l$ is the distance along the path. Hence, entropy production is uniform in the distance along the path. 
	
	We next address the notion of a loss budget, relevant to point (i) above. Indeed, the scale of the loss may appear arbitrary and, if so, one can scale up the loss or the learning rate, such that the implied time-bound goes to zero. Within our continuum description, this is indeed the case, and scaling up the loss would simply speed up the dynamics and scale down the time-bound in a proportional manner. However, as far as our description mimics discrete Gradient Descent (GD), one can only consider small gradients and hence a small loss/learning rate. At higher gradients, discrete GD would start deviating from its continuum approximation, and at even higher learning rates it often leads to NaNs. Analogously to how the binding energy of an atom sets a meaningful energy scale in physics (electron volt), these discrete effects, which depend on model and training choices, set a scale to the loss. The speed limit, as derived from the continuum, implies nothing about this scale. Still, given that we are well below this scale, it bounds the speed of gradient-flow dynamics. In principle, other speed limits relevant to discrete dynamics could be derived based on similar models \cite{Vu_22_06_02684}.
	
	\section{Case studies}
	Here, we present two examples where the speed limit bound can be evaluated analytically. The first example illustrates the interplay between the speed limit, entropy production, and noise in the algorithm for a simple linear perceptron. The second example illustrates how the optimality in training is related to the structure of the spectrum of the NTK, as well as the discrepancy from the target.  
	\subsection{Linear regression - in high dimension}
	Consider the problem of linear regression with scalar output, given a dataset $\mathcal{D}_n=\{X,\bm{y}\}$ where $X\in \mathbb{R}^{d\times n}$, and $\bm{y}\in \mathbb{R}^{n}$. The output of the algorithm is $\hat{y}(\bm{x})=\btheta^{\T}\bm{x}$, where the weights $\btheta\in \mathbb{R}^d$ are 
	learned via Langevin dynamics \eqref{eq:SGD_weight_decay}.
	We consider the squared
	error loss $\cL(\btheta;\cD) =\frac{1}{2}\,\|\bm{y}- X^T\btheta\|^2$ with weight decay with intensity $\lambda d/\beta$.
	In this case, one can provide exact equations for the dynamics of $\mathcal{W}_2(t)$, and $R(t)$ see Appendix \ref{sec:linear_regression_hd} for details of the derivation. 
	
	To gain intuition, below we explore the speed limit bound in the asymptotic regime where the number of samples $n$, commensurate with the input dimension size, $d$, such that, $d/n\to\gamma\in(0,\infty)$,
	while $d,n\to\infty$. To facilitate the analysis, we assume a teacher-student setting with the target model $y=\btheta_\star^\T \bm{x}$, with the true weights $\btheta_\star\sim \mathcal{N}(0,\alpha/d I_d)$. In addition, we assume that
	$X$ had i.i.d. entries. 
	In Appendix \ref{sec:linear_regression_hd} we provide an exact formula for the speed limit bound, $T_{\mathrm{SL}}(\lambda, \beta ,\gamma, \alpha)$, which depends only on these four parameters, noise level, $\beta^{-1}$, the variance of the true weights, $\alpha$, weight decay, $\lambda$, and the limiting dimension ratio, $\gamma$. 
	There are a few interesting limits, one can explore. Taking the limit of $\beta\rightarrow\infty$ i.e. zero noise (gradient descent), the speed limit amount to a specific number, 
	\begin{equation} \label{eq:T_beta_infty_main}T_{\mathrm{SL}}(\beta \to \infty )\to 2\frac{1+\alpha\lambda }{\int s \; d\rho(s)},
	\end{equation}
	with $\rho$ being the Marchenko-Pastor distribution i.e. the limiting eigenvalues' distribution of the covariance matrix $XX^\T/n$. See Appendix \ref{sec:linear_regression_hd} for more details. In this regime, the main source of entropy production in this limit is the loss at initialization.  
	
	On the other hand, if we take the opposite limit of large noise we have, 
	\begin{equation}\label{eq:T_beta_0}
		T_{\mathrm{SL}}({\beta} \to 0) \to 0
	\end{equation} 
	In this regime, the system is driven by noise, and essentially the distribution at the end of training is equal to the distribution at initialization, therefore one can learn at zero time.

	In the large samples' regime,  $n\to\infty$, ($\gamma \to 0$) corresponds to $n\gg d$, the bound reaches the following finite value: 
	\begin{equation} \label{eq:Tsl_n}
		T_{\mathrm{SL}}(n\to \infty) \to 2\lambda \alpha.
	\end{equation} 
	This limit is in essence where we learn the population error. Remarkably, it is independent of the noise level $\beta$. 
	
	Last, in the over-parametrized regime, $d\to \infty$, and $d\gg n$ ($\gamma \to \infty$) we have that
	\begin{equation}\label{eq:T_d_infty}
		T_{\mathrm{SL}}(d \to \infty ) \to 0
	\end{equation}
	Interestingly, in this regime, the parameters are not moving a lot and therefore the final distribution is very close to its initial one. This is not the case when the noise is zero, as follows from  \eqref{eq:T_beta_infty_main}. 
	Therefore, the limit of $d\to \infty$ does not commute with the limit of $\beta\to \infty$. 
	
	We note that \Eqref{eq:Tsl_n} and  \eqref{eq:T_beta_infty_main} show that even in the limit of zero noise and infinite samples, what makes the learning slower is high regularization and high variance of the target true weights.
	
	\subsection{Neural Tangent Kernel (NTK) dynamics}
	As a second analytically tractable example, consider a neural network trained in an NTK setting from a given fixed initial state ($\btheta(0)$) for some time $T$. As no noise is introduced, $T$ determines the final state ($\btheta(T)$) and decrease of the loss ($\Delta L(T)$). Consequently, one can think of the time-bound here as a function of $T$. We define inefficiency via the ratio $T/T_{\text{SL}}(T) \geq 1$. Specifically, it is given by 
	\begin{align}
		\frac{T}{T_{\text{SL}}(T)} &= \frac{T \Delta L(T)}{|\btheta(0)-\btheta(T)|^2}
	\end{align}
	
	The NTK dynamics, being linear, lends itself to exact analytical expressions for all quantities involved. Specifically, 
	\begin{align}
		|\btheta(0)-\btheta(T)|^2 &= \sum_{\lambda} \Delta_{\lambda}^2 \, \lambda^{-1} \left[1-e^{-\lambda T} \right]^2\\ \nonumber 
		\Delta L(T) &=  \sum_{\lambda}\Delta_{\lambda}^2 \left[1-e^{-2\lambda T}\right],
	\end{align}
	where the summation is over all NTK train kernel eigenvalues and $\Delta_{\lambda}$ is the difference between the network's train outputs at initialization and the target projected on the eigenvector associated with $\lambda$. 
	
	Making several experimentally motivated assumptions on $\Delta_{\lambda}$ and $\lambda$ we next derive concrete asymptotic results for the inefficiency ratio. Specifically, we assume $\lambda_{k} = k^{-\alpha}$ and $\Delta^2_{\lambda_k} = k^{-\delta}$ where $k \in 1, \ldots, n$. Assuming $T \propto \lambda_{n}^{-1}$ such that the lowest mode is partially learned, as well as  $\alpha,\delta>0$, and $0<\alpha^{-1}(1-\delta)<1$ we find the following large $T$ asymptotic
	\begin{align}
		\mathcal{W}_2 &= \sum_{\lambda} \frac{\Delta_{\lambda}^2[1-e^{-\lambda T}]^2}{\lambda} \propto T^{\alpha^{-1}(1-\delta)+1} \\ \nonumber
		\beta^{-1}R &= \sum_{\lambda}\Delta_{\lambda}^2\left[1-e^{-2\lambda t}\right] \propto T^{\alpha^{-1}(1-\delta)}
	\end{align}
	whereas for $-1<\alpha^{-1}(1-\delta)<0$ we find 
	\begin{align}
		\beta^{-1}R &\propto T^{0},
	\end{align}
	and $\mathcal{W}_2$ remains with the same scaling. Remarkably, in the first regime, we find  $T_{\text{SL}}(T) \propto T$. Since the proportionality factors are all $O(1)$, we thus find an optimal behavior in the scaling sense. In contrast, for, $\delta>1$ we enter the second regime leading to $T_{\text{SL}}(T) \propto T^{1+\alpha^{-1}(1-\delta)}$. Noting that the exponent is now smaller than, $1$ we obtain a non-optimal behavior in the scaling sense.
	
	Interestingly, if the target is small compared to the outputs at initialization, $\Delta_{\lambda}^2$ would be dominated by the output of the network at initialization which is given by a random draw from the NNGP. If, based on their similar performance, we ignore differences between the NTK and NNGP spectra, we have that the discrepancy $\Delta_{\lambda}^2$ scales as $\lambda$. Furthermore, we  observe $\alpha=1$, which means that if $f$ dominated, the residue we have $\delta=\alpha=1$, placing us exactly at the threshold value between the efficient and inefficient regime.  
	
	{\bf Geometric aspects.} Next, we explore some geometrical aspects of the dynamics, namely how different the length $l_{\gamma}$ of the curve traveled in weight space is compared to the length $l_{\text{geo}}=\sqrt{\mathcal{W}_2}$ of the optimal path, which is a straight line. As shown in \prettyref{app:NTK} the length of both curves as a function $T$ scales identically   
	\begin{align}
		l_{\text{geo}}(T) &\propto T^{(\alpha^{-1}+1-\delta/\alpha)/2} \\ \nonumber 
		l_{\gamma}(T) &\propto  T^{(\alpha^{-1}+1-\delta/\alpha)/2}
	\end{align}
	Interestingly, we find the same asymptotic for the lengths, independent of $\alpha$ and $\delta$ (for  $\alpha>0,\delta\ge 0$ and $(\alpha^{-1}+1-\delta/\alpha)>0$). This means that at least within this NTK limit, inefficiency is not attributed to having a highly twisted and long curve, but rather having highly inhomogeneous velocity along the curve. 
	
	\section{Experiments on CIFAR-10}
	Here, we study the efficiency of simple CNNs trained on real-world data. Specifically, we train Myrtle-5, a 5 trainable-layers convolutional network with several pooling layers, having 128 channels on subsets of CIFAR-10 with up to $5k$ samples. Training is carried out for 200k epochs using MSE loss, full batch gradient descent, and small learning rates ($10^{-4}$ to $10^{-5}$) to assure closeness to gradient flow. We train 6 realizations of such networks, with different initialization seeds, and use datasets consisting of the first $n=500,1250,2500,5000$ samples of CIFAR-10. We record the gradients, losses, and network weights along the path. These enable us to estimate the Wasserstein-2 distance ($L_2$ distance in weight-space) and entropy production (drop in loss) for each realization as a function of time. From these, we obtain the inefficiency ratio per-realization ($T/T_{\text{SL}}(T)$) and geometric inefficiency ratio ($l_{\text{NTK}}(T)/l_{\text{geo}}(T)$). We furthermore obtain the empirical NTK spectrum and the overlap between initialization residues and the NTK eigenvectors. 
	
\begin{figure}
		\centering   
		\includegraphics[width=\columnwidth]{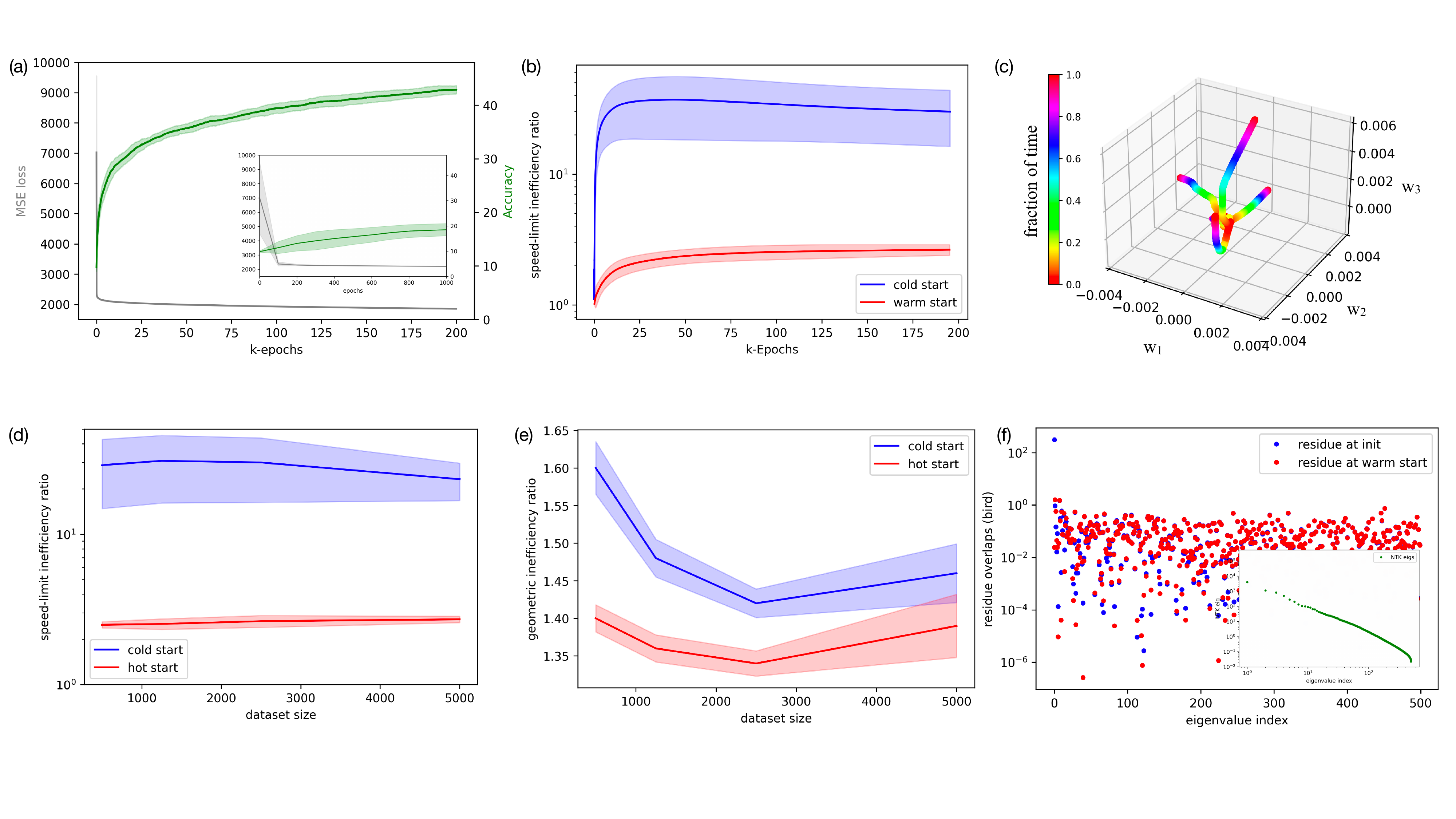}
		\caption{{{\bf Efficiency aspects of Myrtle-5 CNN trained on CIFAR-10.} Panel (a): MSE loss and test accuracy for six networks trained on $2500$ data points. A dramatic initial decrease in loss is evident without similar improvement in accuracy. Shaded areas reflect standard deviations across networks. Panel (b): For the same networks, speed limit as a function of epoch w.r.t. initialization (cold start) or epoch $2000$ (warm start). Most of the inefficiency is thus attributed to the fast entropy burn near initialization. Panel (c): Again for $n=2500$, dynamics of 6 randomly chosen weight-triplets ($w_1,w_2,w_3$). Panel (d): Inefficiency ratio at epoch 200k as a function of dataset size. Panel (e): Ratio of the curve length traveled during training over the optimal curve, again for different data-set sizes. Panel (f): Overlap of residue at initialization and at the warm start with the NTK eigenvectors. Most of the entropy burn can be associated with the first few eigenvalues, which are quickly learned and hence removed from the residue. Inset: NTK eigenvalues based on a single network with $n=500$ datapoints. } 
			\label{fig:MyrtleSpeedLimit}}
	\end{figure}

	As shown in Fig. \ref{fig:MyrtleSpeedLimit}, the very early stages of the dynamics are associated with a fast increase in entropy ($\beta^{-1} R$) or, equivalently, a drop in MSE loss. However, the accuracy does not show any marked features during this process. This motivates us to explore two notions of inefficiency, one measured with respect to the network's initialization (cold start) and the other with respect to the first time at which test accuracy averaged over realizations reached 12\% (warm start). We note that for $n=500$ we reach a final test accuracy of, $30\% \pm 1\%$ whereas for $n=5000$ we obtain $47\% \pm 1\%$.
	
	Our main results are given in Fig. \ref{fig:MyrtleSpeedLimit}. These support the following rather unexpected picture. Apart from an initial stage at which few very high NTK kernel eigenvalue are learned, the dynamics of this real-world network trained on real-world data seems optimal up to, a roughly constant, $O(1)$ factor. While in principle, one could have expected factors proportional to dataset size or training time, these seem to cancel out. 
	
	Similarly, the length of the curve traveled during training in weight space coincides with the $L_2$ length up to a $O(1)$ factor (panel (e)). Panel (c) further tracks several different 3d projections of the path traveled in weight-space (namely the curve $(w_1(t),w_2(t),w_3(t))$ where $w_i$ are some randomly chosen subset of $\btheta$) showing rather few twists and turns. 
	
	Though the actual NTK kernel of this network is not constant during training, these results are in qualitative agreement with the theoretical results given in the NTK section, where it was assumed constant.  
	
	\section{Discussion}
	In this work, we set out to explore learning dynamics in deep neural networks from a thermodynamic standpoint. We fleshed out how several key concepts in thermodynamics, such as entropy production and the thermodynamic speed limit, carry through to the deep learning realm. Analytical formulas for these quantities were derived for two simple models, a linear perceptron and a network trained in the NTK regime. Interestingly, following some realistic scaling assumptions on the NTK spectrum over-parameterized neural networks trained with gradient flow revealed surprising efficiency, leaving only $O(1)$ improvement factors to be desired. Similarly, distance-wise, the curved traveled in weight space during training does not differ much from a straight line. Our theoretical results were supported by small-scale experiments on convolutional networks trained in CIFAR-10.    
	
	Various aspects of this work invite further study. It would be interesting to extend our theory to finite learning rates so that it can include discretization effects. This would also shed light on what are the allowed entropy budgets, thereby setting a definite scale for the time-bound. Extending our results to finite-width neutral networks, perhaps using kernel-adaptation methods \cite{seroussi2023separation,LiSompolinsky2021,ariosto2022statistical,bordelon2023dynamics}, would enable us to study the thermodynamic implications of feature-learning effects. Finally, it is desirable to extend our experiments to a wider range of networks and see what practical improvements to training can be gained from this physical viewpoint.  
	
	\bibliographystyle{plain}
	\bibliography{refs}

\begin{thebibliography}{10}

\bibitem{ariosto2022statistical}
S~Ariosto, R~Pacelli, M~Pastore, F~Ginelli, M~Gherardi, and P~Rotondo.
\newblock Statistical mechanics of deep learning beyond the infinite-width
  limit.
\newblock {\em arXiv preprint arXiv:2209.04882}, 2022.

\bibitem{Benamou00_375}
JD~Benamou and Y.~Brenier.
\newblock A computational fluid mechanics solution to the monge-kantorovich
  mass transfer problem.
\newblock {\em Numer. Math.}, 84:375--393, 2000.

\bibitem{bordelon2023dynamics}
Blake Bordelon and Cengiz Pehlevan.
\newblock Dynamics of finite width kernel and prediction fluctuations in mean
  field neural networks.
\newblock {\em arXiv preprint arXiv:2304.03408}, 2023.

\bibitem{Crooks1999_2721}
Gavin~E. Crooks.
\newblock Entropy production fluctuation theorem and the nonequilibrium work
  relation for free energy differences.
\newblock {\em Physical Review E}, 60(3):2721--2726, September 1999.

\bibitem{cuesta1996lower}
Juan~Antonio Cuesta-Albertos, Carlos Matr{\'a}n-Bea, and Araceli Tuero-Diaz.
\newblock On lower bounds for the l 2-wasserstein metric in a hilbert space.
\newblock {\em Journal of Theoretical Probability}, 9(2):263--283, 1996.

\bibitem{Helias20_970}
Moritz Helias and David Dahmen.
\newblock {\em Statistical Field Theory for Neural Networks}.
\newblock Springer International Publishing, 2020.

\bibitem{jacot2018}
Arthur {Jacot}, Franck {Gabriel}, and Cl{\'e}ment {Hongler}.
\newblock {Neural Tangent Kernel: Convergence and Generalization in Neural
  Networks}.
\newblock {\em arXiv e-prints}, page arXiv:1806.07572, Jun 2018.

\bibitem{LiSompolinsky2021}
Qianyi Li and Haim Sompolinsky.
\newblock Statistical mechanics of deep linear neural networks: The
  backpropagating kernel renormalization.
\newblock {\em Phys. Rev. X}, 11:031059, Sep 2021.

\bibitem{nguyen2009surrogate}
XuanLong Nguyen, Martin~J Wainwright, and Michael~I Jordan.
\newblock On surrogate loss functions and f-divergences.
\newblock 2009.

\bibitem{Onsager53}
L.~Onsager and S.~Machlup.
\newblock Fluctuations and irreversible processes.
\newblock 91:1505--1512, Sep 1953.

\bibitem{Risken96}
Hannes Risken.
\newblock {\em The Fokker-Planck Equation}.
\newblock Springer Verlag Berlin Heidelberg, 1996.

\bibitem{Seifert12_126001}
Udo Seifert.
\newblock Stochastic thermodynamics, fluctuation theorems, and molecular
  machines.
\newblock 75:126001, 2012.

\bibitem{seroussi2023separation}
Inbar Seroussi, Gadi Naveh, and Zohar Ringel.
\newblock Separation of scales and a thermodynamic description of feature
  learning in some cnns.
\newblock {\em Nature Communications}, 14(1):908, 2023.

\bibitem{Vu_22_06_02684}
Tan~Van Vu and Keiji Saito.
\newblock Thermodynamic unification of optimal transport: Thermodynamic
  uncertainty relation, minimum dissipation, and thermodynamic speed limits.
\newblock 2022.

\bibitem{yuan2021large}
Zhuoning Yuan, Yan Yan, Milan Sonka, and Tianbao Yang.
\newblock Large-scale robust deep auc maximization: A new surrogate loss and
  empirical studies on medical image classification.
\newblock In {\em Proceedings of the IEEE/CVF International Conference on
  Computer Vision}, pages 3040--3049, 2021.

\end{thebibliography}
	
	\appendix
	
	\section{Conditional distribution, reversal of time and entropy production}
	In this section, we provide a derivation of \eqref{eq:R_learning}, and \eqref{eq:R_geom_loss} in the main text without loss of generality we take $\eta = 1$. We then show that these two definitions are consistent.  
	
	\subsection{Entropy production at Equilibrium }
	Similar to, \cite{Seifert12_126001} we split the definition of the entropy production in \eqref{eq:def_R} into two parts: 
	
	\begin{align}
		R & =R_{0}+R_{1},\label{eq:def_R}\\
		R_{0} & =\Big\langle\ln\,\frac{p(\btheta(0))}{p(\btheta(T))}\Big\rangle\qquad R_{1}=\Big\langle\ln\,\frac{p(\btheta(T)|\btheta(0))}{q(\btheta(0)|\btheta(T))}\Big\rangle.\nonumber 
	\end{align}
	The initial and final distributions, adopting statistical physics notation, are
	\begin{align}
		p_{0}(\btheta) & =\cZ_{0}^{-1}\,e^{-\beta\|\btheta\|^{2}}\label{eq:initialization}
	\end{align}
	with normalization $\cZ_{0}=(\pi\,\beta^{-1}\big)^{\frac{P}{2}}$ known as the partition function.
	The temperature determines the variance $(2\beta)^{-1}$ of this Gaussian
	initial distribution of the weights.
	
	At time $T$ the stationary distribution of the weights is 
	\begin{align}
		p_{T}(\btheta) & =\cZ_{T}^{-1}\,e^{-\beta\,(\|\btheta\|^{2}+\cL(\btheta; \mathcal{D}))},\label{eq:final_distribution}
	\end{align}
	where $\cZ_{T}=\int\,e^{-\beta\,(\|\btheta\|^{2}+\cL(\btheta;\mathcal{D}))}\,d\btheta$
	is the normalization.
	The first term, with \eqref{eq:initialization} and \eqref{eq:final_distribution},
	yields
	\begin{align}
		R_{0} & =\beta\,\langle \|\btheta(T)\|^{2}+\cL(\btheta(T);\mathcal{D})\rangle-\beta\,\langle \|\btheta(0)\|^{2}\rangle\label{eq:R_0}\\
		& +\ln\cZ_{T}-\ln\cZ_{0}.\nonumber 
	\end{align}
	The second term $R_{1}$ measures the log ratio of the process running
	forward versus backward. For the conservative force in \prettyref{eq:SGD_weight_decay}
	it can be shown (see \prettyref{sec:Irreversibility}, \prettyref{eq:R_1_path})
	to take the value
	\begin{align*}
		R_{1} & =\beta\,\langle \cL(\btheta(0);\mathcal{D})+\|\btheta(0)\|^{2}\rangle-\beta\,\langle \|\btheta(T)\|^{2}+\cL(\btheta(T);\mathcal{D})\rangle.
	\end{align*}
	So in total with $V$ expressed by \prettyref{eq:switch_potential}
	we get the irreversibility
	\begin{align}
		\boxed{R=\ln\cZ_{T}-\ln\cZ_{0}+\beta\,\langle\cL(\btheta(0))\rangle}.\label{eq:R_learning}
	\end{align}
	This result expresses the irreversibility
	of the learning process in terms of equilibrium properties, the free
	energies of the weight distribution at initialization $\ln\cZ_{0}$
	and after learning $\ln\cZ_{T}$ and the expected initial loss.
	
	\subsection{Entropy production from dynamics} \label{app:entropy_from_dyn}
	
	Likewise, entropy production \prettyref{eq:def_R} can be expressed
	as a dynamical quantity \cite{Vu_22_06_02684}, in terms of
	the stochastic velocity field $\bv(\btheta,t)$ (for details see \prettyref{app:Fokker-Planck-equilibrium},
	i.p. \prettyref{eq:velocity}), which turns the Fokker-Planck equation
	for the temporal evolution of the density $p(\btheta,t)$ into an
	effective transport equation, 
	\begin{align}
		\partial_{t}\,p(\btheta,t)+\nabla_{\theta}\cdot\big[\bv(\btheta,t)\,p(\btheta,t)\big] & =0.\label{eq:FP_as_transport}
	\end{align}
	Here $\bv$ can be thought of as an effective deterministic velocity
	field that would cause the same evolution of $p(\btheta,t)$ as does
	the stochastic process \prettyref{eq:SGD_weight_decay}. Entropy production
	then takes the form (\prettyref{app:Equivalence-of-entropies})
	\begin{align}
		R=\beta\,\int_{0}^{T}\,\langle\|\bv(\btheta,t)\|^{2}\rangle\,dt,\label{eq:entropy_rate_main_text}
	\end{align}
	which, in the case of a conservative force of the learning dynamics
	(cf. \prettyref{eq:R_Vu_conservative}), reads
	\begin{align}
		\boxed{R=\int_{0}^{T}\,\beta\,\langle\|\nabla_{\btheta}V\|^{2}\rangle-2\,\langle\Delta_{\btheta}V\rangle+\beta^{-1}\,\langle\|\nabla_{\btheta}\ln p\|^{2}\rangle\,dt}.\label{eq:R_geom_loss_app}
	\end{align}
	In the low noise limit $\beta\gg1$ the first term $\propto\beta^{1}$
	dominates, which has the simple interpretation of the average squared
	length of the gradient. The next to this leading term is $\propto\beta^{0}$
	which contains the average Hessian of the loss function. Equating
	\prettyref{eq:R_geom_loss} and \prettyref{eq:R_learning} therefore
	relates the geometry of the loss landscape to equilibrium properties
	of the initial and the final distribution of the weights. This is
	the second theoretical result of this work.
	
	\subsection{Speed limits from optimal transport} \label{app:speed_limit}
	
	The stochastic velocity for $\bv$ appearing in \prettyref{eq:entropy_rate_main_text}
	is key to linking entropy production to the distance between the initial
	and final distribution of the weights and to optimal transport. This
	velocity enables the definition of a measure of the distance between
	two probability distributions, the Wasserstein-2-distance, \cite[their Eq. (11)]{Vu_22_06_02684}
	\begin{align}
		\mathcal{W}_{2}(p_{0},p_{T}) & :=\min_{\bv}\,T\,\int_{0}^{T}\langle\|\bv(\btheta,t)\|^{2}\rangle\,dt,\label{eq:W_2_main}
	\end{align}
	where minimization is performed under the constraint that the velocity
	field $\bv(\btheta,0\le t\le T)$ transforms $p_{0}(\btheta)$ into
	$p_{T}(\btheta)$ by \prettyref{eq:FP_as_transport}. The right-hand
	side of \prettyref{eq:W_2_main} contains the time-averaged mean squared
	velocity $\langle\|\bv(\btheta,t)\|^{2}\rangle$ required for optimal
	transport. Comparing \prettyref{eq:W_2_main} to \prettyref{eq:entropy_rate_main_text},
	the learning process is but one possible transport solution, not necessarily
	the optimal one though, so one obtains the thermodynamic speed limit
	known as the Benamou--Brenier formula \cite{Benamou00_375,Vu_22_06_02684}
	
	\begin{align}
		\boxed{T\ge\frac{\beta\,\mathcal{W}_2(p_{0},p_{T})}{R}} & ,
	\end{align}
	which is the third theoretical relation to be explored in the following.
	It provides a lower bound on the time $T$ for a stochastic process
	to evolve $p_{0}$ into $p_{T}$, which depends on the distance $\mathcal{W}_2$
	between the two distributions $p_{0}$ and $p_{T}$ and on the amount
	of entropy $R$ produced at a given temperature $\beta$.
	
	\subsection{Path measure}
	
	We here follow \cite[i.p. Sec 4]{Seifert12_126001} and \cite[i.p. Sec 7.2]{Helias20_970}.
	Assuming It\^{o} convention, 
	the stochastic differential equation (SDE) \eqref{eq:SGD_weight_decay} needs to be evaluated in discrete time, as
	\begin{align}
		\btheta(t+dt) & =\btheta(t)-\nabla_{\btheta} V(\btheta(t);\mathcal{D}) \,dt+d\cB(t)   ,\label{eq:Ito}\\
		d\cB(t)_{i} & \stackrel{\text{i.i.d.}}{\sim}\cN(0,2\beta^{-1}I_P\,dt).\nonumber 
	\end{align}
	The important point here is that the drift
	is evaluated at the left boundary $t$ of any time interval $[t,t+dt]$.
	The dynamics \eqref{eq:SGD_weight_decay} implies a measure on the path $\btheta(t)$
	for $t\in[0,T]$. In the following consider discretized time, introducing
	the temporal indices $l$ as $\btheta_{l}:=\btheta(l\,dt)$, $d\cB_{l}:=d\cB(l\,dt)$,
	and $\bbf_{l}:=\bbf(\btheta_{l},l\,dt) = \nabla_{\btheta_l} V(\btheta_l;\mathcal{D}) $. In this notation, the Ito update
	step in \eqref{eq:Ito} takes the form
	\begin{align}
		\btheta_{l+1} & =\btheta_{l}+\bbf_{l}\,dt+d\cB_{l}\qquad0\le l\le T/dt,\label{eq:discrete_update}\\
		\btheta_{0} & =\btheta(0).\nonumber 
	\end{align}
	The measure on the path $\btheta_{1},\ldots,\btheta_{T/dt}$ is induced by
	the Gaussian measure $\propto\exp\big(-\frac{\beta}{4 dt}\,\sum_{i=0}^{T/dt}\,\|d\cB_{l}\|^{2}\big)$
	of the stochastic increments $d\cB_{l,i}\stackrel{\text{i.i.d.}}{\sim}\cN(0,2\beta^{-1} dt)$.
	Solving \eqref{eq:discrete_update} for $d\mathcal{B}_{l}=\btheta_{l+1}-\btheta_{l}-\bbf_{l}\,dt$
	one has
	
	\begin{align}
		p(\btheta_{1},\ldots,\btheta_{T/dt}|\btheta_{0}) & \propto\exp\big(-\frac{\beta}{4}\,\sum_{i=0}^{T/dt}\left\Vert \frac{\btheta_{l+1}-\btheta_{l}}{dt}-\bbf_{l}\right\Vert ^{2}\,dt\big).\label{eq:discrete_path_measure}
	\end{align}
	Symbolically, one may therefore write the measure on the path $\btheta(0\le t\le T)$
	as a functional
	\begin{align}
		p[\btheta(0\le t\le T)] & \propto\exp\big(\int_{0}^{T}\,A[\btheta](t)\,dt\big),\label{eq:path_measure}
	\end{align}
	where $A$ denotes the time-local Lagrangian (also known as the Onsager-Machlup
	action \cite{Onsager53}, reviewed in \cite[i.p. Sec 7.2]{Helias20_970})
	\begin{align}
		A[\btheta](t) & =-\frac{\beta}{4}\,\big[\partial_{t}\btheta(t)-\bbf(\btheta(t),t)\big]^{2}.\label{eq:Lagrangian}
	\end{align}
	Note, however, that in the symbolic notation the Ito procedure as
	well as the initial condition are both implicit.
	
	\subsection{Fokker-Planck equation and equilibrium distribution\label{app:Fokker-Planck-equilibrium}}
	
	The above process can also be represented in terms of macroscopic quantities, such as probability density. The probability density satisfies the Fokker-Planck equation.  This equation takes 
	the form of a continuity equation (cf. \cite{Risken96})
	\begin{align}
		\partial_{t}\,p(\btheta,t) & =-\nabla_{\btheta}\cdot\bJ\,(\btheta,t),\label{eq:FP_orig}
	\end{align}
	with the probability current $\bJ$
	\begin{align}
		J(\btheta,t) & =\big(\bbf(\btheta,t)-\beta^{-1}\,\nabla_{\btheta}\big)\,p(\btheta,t).\label{eq:prob_current}
	\end{align}
	For a conservative force $\bbf(\btheta)=-\nabla_{\btheta}V(\btheta)$ the stationary
	distribution is of Boltzmann form
	\begin{align}
		p_{0}(\btheta) & \propto e^{-\beta V(\btheta)},\label{eq:p_equilibrium}
	\end{align}
	for which the probability current $J(\btheta)\equiv0$ vanishes. A different
	way of writing the Fokker-Planck equation \prettyref{eq:FP_orig}
	is in the form of a transport equation where the probability current
	$\bJ=\bv\,p$ is the product of velocity $\bv$ and probability $p$,
	namely
	\begin{align}
		\partial_{t}\,p(\btheta,t) & =-\nabla_{\btheta}\cdot\big[\bv(\btheta,t)\,p(\btheta,t)\big],\label{eq:FP_second_notation}\\
		\bv(\btheta,t) & =\bbf(\btheta)-\beta^{-1}\,\nabla_{\btheta}\,\ln\,p(\btheta,t).\label{eq:velocity}
	\end{align}
	The additional term $-\beta^{-1}\,\nabla_{\btheta}\ln p$ can be regarded
	as an entropic force. The interpretation of $\bv$ as a velocity makes
	sense, because it may be interpreted as the probability current $\bJ$
	conditioned on finding the system in state $\btheta$ at time $t$. For
	a system in thermodynamic equilibrium \prettyref{eq:p_equilibrium}, the
	velocity vanishes at each point $\btheta$, because $\bv_{0}(\btheta)=\bJ_{0}(\btheta)/p_{0}(\btheta)\equiv0$.
	\subsection{Irreversibility with conservative forces\label{sec:Irreversibility}}
	
	To measure the irreversibility, we need the ratio of probabilities
	\prettyref{eq:def_R}
	\begin{align}
		R_{1}=\Big\langle\ln\,\frac{p(\btheta(T)|\btheta(0))}{q(\btheta(0)|\btheta(T))}\Big\rangle & ,\label{eq:R_1}
	\end{align}
	where $p$ denotes the measure \prettyref{eq:path_measure} on the
	path $\btheta$ running forward in time and $q$ denotes the probability
	assigned to a path $\tilde{\btheta}$ by the measure \prettyref{eq:path_measure}
	if one reverses the temporal sequence of state traversals
	\begin{align*}
		\tilde{\btheta}(t) & :=\btheta(T-t)\quad0\le t\le T.
	\end{align*}
	The reversed path $\tilde{\btheta}$ is constructed such that its initial
	point $\tilde{\btheta}(0)$ is identical to the final point of the forward
	dynamics $\btheta(T)$, so $\tilde{\btheta}(0)=\btheta(T)$. The average in \prettyref{eq:R_1}
	is over the ensemble of all paths that started at $t=-\infty$, thus
	it is identical to the expectation over all random initializations at $t=0$.
	
	Inserting $\tilde{\btheta}$ into the Lagrangian $A$ \prettyref{eq:Lagrangian}
	only the mixed term $\beta/2\,\bbf(\btheta(t))\cdot\partial_{t}\btheta(t)$
	changes sign, so that \eqref{eq:R_1} reads
	\begin{align}
		R_{1} & =\beta\,\Big\langle\int_{0}^{T}\big[\bbf(\btheta(t))\cdot\partial_{t}\,\btheta(t)\big]\,dt\Big\rangle\nonumber \\
		& =\beta\,\Big\langle\int_{\btheta(0)}^{\btheta(T)}\bbf(\btheta)\cdot d\btheta\Big\rangle\nonumber \\
		& =\beta\,(\big\langle V(\btheta(0);\mathcal{D})\big\rangle-\big\langle V(\btheta(T), \mathcal{D})\big\rangle),\label{eq:R_1_path}
	\end{align}
	where the penultimate line holds for any non-equilibrium Langevin
	dynamics with time-independent force $\bbf(\btheta(t))$ and the last
	line holds in case that $\bbf$ is conservative. In the latter case,
	irreversibility depends linearly on the difference in energy $\Delta V$
	between initial and final state. Physically, this is the work that
	the heat bath has exerted on the system \cite[i.p. their Eq. (6)]{Crooks1999_2721}.
	The irreversibility $R$ defined in \prettyref{eq:def_R} for a conservative
	force $\bbf=-\nabla_{\btheta}V$ thus is, 
	\begin{align}
		R & =\langle\ln\,p(\btheta(0))\rangle-\langle\ln\,p(\btheta(T))\rangle\label{eq:R_final_crooks}\\
		& +\beta\,\big\langle V(\btheta(0))\big\rangle-\beta\,\big\langle V(\btheta(T))\big\rangle,\nonumber 
	\end{align}
	which corresponds to Eq. (6) in \cite{Crooks1999_2721}.

	\subsection{Irreversibility from stochastic velocity\label{app:Equivalence-of-entropies}}
	
	We here show that, in the case of conservative forces, the irreversibility
	\prettyref{eq:R_final_crooks} obtained from the initial and final
	equilibrium distribution is identical to the dynamic expression \prettyref{eq:entropy_rate_main_text}.
	To show the equivalence, consider the temporal change of the mean
	of any observable $O(\btheta)$ is $\partial_{t}\langle O\rangle=\partial_{t}\,\int_{\Omega}\,p(\btheta,t)\,O(\btheta)\,d\btheta$.
	Choosing in particular $V$ as the observable $O=V$ the temporal
	change of the potential is
	\begin{align}
		\partial_{t}\,\langle V(\btheta(t))\rangle & =\int_{\Omega}\,V(\btheta(t))\,\partial_{t}\,p(\btheta,t)\,d\btheta\label{eq:dV_dt}\\
		& \stackrel{(\ref{eq:FP_second_notation})}{=}-\int_{\Omega}\,V(\btheta)\,\nabla_{\btheta}\cdot\big[\bv(\btheta,t)\,p(\btheta,t)\big]\,d\btheta\nonumber \\
		& \stackrel{\text{i.b.p.}}{=}\int_{\Omega}\,\big[\nabla_{\btheta}V(\btheta)\big]\cdot\bv(\bx,t)\,p(\btheta,t)\,d\btheta\nonumber \\
		& =-\int_{\Omega}\,\bbf(\btheta)\cdot\bv(\btheta,t)\,p(\btheta,t)\,d\btheta,\nonumber 
	\end{align}
	where we assumed that $p(\btheta,t)$ declines sufficiently quickly with
	$\|\btheta\|\to\infty$, so boundary terms vanish when integrating by parts (i.b.p.). So we find
	\begin{align*}
		& \int_{\Omega}\,\|\bv(\btheta,t)\|^{2}\,p(\btheta,t)\,d\btheta\\
		\stackrel{(\ref{eq:velocity})}{=} & \int_{\Omega}\,\big[\bbf(\btheta)-\beta^{-1}\,\nabla_{\btheta}\ln p(\btheta,t)\big]\cdot\bv(\btheta,t)\,p(\btheta,t)\,d\btheta\\
		\stackrel{(\ref{eq:dV_dt})}{=} & -\partial_{t}\langle V(\bx(t))\rangle-\beta^{-1}\,\int_{\Omega}\,\big[\nabla_{\btheta}\ln p(\btheta,t)\big]\cdot\bv(\bx,t)\,p(\btheta,t)\,d\btheta.
	\end{align*}
	Integration by parts of the latter integral, again using vanishing
	boundary terms for $\|\btheta\|\to\infty$, it is
	\begin{align}
		& -\int_{\Omega}\,\big[\nabla_{\btheta}\ln p(\btheta,t)\big]\cdot\bv(\btheta,t)\,p(\btheta,t)\,d\btheta\label{eq:temp_diff}\\
		= & \int_{\Omega}\,\ln p(\btheta,t)\,\nabla_{\btheta}\cdot\big[\bv(\btheta,t)\,p(\btheta,t)\big]\,d\btheta\nonumber \\
		\stackrel{(\ref{eq:FP_second_notation})}{=} & -\int_{\Omega}\,\ln p(\btheta,t)\,\partial_{t}\,p(\btheta,t)\,d\btheta.\nonumber 
	\end{align}
	The latter integral is identical to
	\begin{align}
		& -\partial_{t}\,\int_{\Omega}\,\ln p(\btheta,t)\,p(\btheta,t)\,d\btheta\label{eq:temp_diff_2}\\
		= & -\underbrace{\partial_{t}\,\int_{\Omega}\,p(\btheta,t)\,d\btheta}_{=0}-\int_{\Omega}\,\ln p(\btheta,t)\,\partial_{t}\,p(\btheta,t)\,d\btheta.\nonumber 
	\end{align}
	So together we find the differential form of \prettyref{eq:entropy_rate_main_text}
	
	\begin{align}
		\int_{\Omega}\,\|\bv(\btheta,t)\|^{2}\,p(\btheta,t)\,d\btheta & =-\partial_{t}\,\langle V(\btheta(t))\rangle-\beta^{-1}\,\partial_{t}\,\int_{\Omega}\,\ln p(\btheta,t)\,p(\btheta,t)\,d\btheta.\label{eq:dR_dt}
	\end{align}
	Taking the temporal integral over the interval $[0,T]$ we arrive
	at
	\begin{align}
		\beta\,\int_{0}^{T}\,\int_{\Omega}\,\|\bv(\btheta,t)||^{2}\,p(\btheta,t)\,d\btheta\,dt & =\beta\,\langle V(\btheta(0))\rangle-\beta\,\langle V(\btheta(T))\rangle\label{eq:int_v2_final}\\
		& +\langle\ln p(\btheta(0)\rangle-\langle\ln p(\btheta(T))\rangle\nonumber \\
		& \stackrel{(\ref{eq:R_final_crooks})}{=}R.\nonumber 
	\end{align}
	The last line is the difference in the entropy between the initial and
	final state and the right-hand side is identical to \prettyref{eq:R_final_crooks}.
	
	The differential form \eqref{eq:dR_dt}, rewritten more briefly as,
	\begin{align}
		\langle\|\bv(\btheta,t)\|^{2}\rangle & =-\partial_{t}\,\big(\langle V(\btheta(t))\rangle+\beta^{-1}\,\langle\ln\,p(\btheta,t)\rangle\big)\label{eq:average_length_change_free_energy}
	\end{align}
	has an interesting interpretation. In equilibrium statistical mechanics
	one has $p(\btheta)=\cZ^{-1}\,e^{-\beta V(\btheta)}$ which, taking
	the $\ln$ and then the expectation value over $p(\btheta)$, yields
	the usual relation
	\begin{align}
		F:=-\beta^{-1}\,\ln\,\cZ & =\langle V(\btheta)\rangle+\beta^{-1}\,\langle\ln\,p(\btheta)\rangle.\label{eq:free_energy}
	\end{align}
	between free energy $F$, inner energy $\langle V(\btheta)\rangle$,
	and entropy $S=-k_{B}\,\langle\ln\,p(\btheta)\rangle$.
	
	So comparing the right-hand sides \eqref{eq:average_length_change_free_energy}
	and \eqref{eq:free_energy} and defining a ``time-dependent free
	energy'' $F(t):=\langle V(\btheta(t))\rangle+\beta^{-1}\,\langle\ln\,p(\btheta(t))\rangle$,
	one has
	\begin{align*}
		\partial_{t}F(t) & =-\langle\|\bv(\btheta,t)\|^{2}\rangle,
	\end{align*}
	which, by the non-negativity of the right-hand side, shows that $F(t)$
	is a non-increasing function under the Langevin dynamics. Integrated
	over time, $t\in[0,T]$ this yields
	\begin{align*}
		\Delta F=F(T)-F(0) & =-\int_{0}^{T}\,\langle\|\bv(\btheta,t)\|^{2}\rangle\,dt=-\beta^{-1}\,R.
	\end{align*}
	
	Using the above formula for $R$ as a function of the velocity field $\bv$ and the Fokker-Planck equation for the equilibrium density, $p$,
	
	\begin{align}\label{eq:R_Vu}
		R= & \beta\,\int_{0}^{T}\int\,||\bv(\btheta,t)||^{2}\,p(\btheta,t)\,d\btheta\,dt\\
		= & \beta\,\int_{0}^{T}\int\,\big\|\frac{\bJ\,p(\btheta,t)}{p(\btheta,t)}\|^{2}\,p(\btheta,t)\,d\btheta\,dt\nonumber \\
		= & \beta\,\int_{0}^{T}\int\,\frac{\big\|\bJ\,p(\btheta,t)\big\|^{2}}{p(\btheta,t)}\,d\btheta\,dt\nonumber \\
		= & \beta\,\int_{0}^{T}\int\,\frac{\big\|(\bbf(\btheta)-\beta^{-1}\,\nabla_{\btheta})\,p(\btheta,t)\big\|^{2}}{p(\btheta,t)}\,d\btheta\,dt\nonumber \\
		= & \beta\,\int_{0}^{T}\int\,||\bbf(\btheta)||^{2}\,p(\btheta,t)-2\beta^{-1}\,\bbf(\btheta)\cdot\nabla_{\btheta}p(\btheta,t)+\beta^{-2}\,\frac{\nabla_{\btheta}p(\btheta,t)\cdot\nabla_{\btheta}p(\btheta,t)}{p(\btheta,t)}\,d\btheta\,dt\nonumber \\
		= & \int_{0}^{T}\,\beta\,\langle||\bbf(\btheta(t))||^{2}\rangle+2\beta^{-1}\,\langle\nabla_{\btheta}\cdot \bbf\rangle+\beta^{-1}\,\langle||\nabla_{\btheta}\ln\,p||^{2}\rangle\,dt \nonumber, 
	\end{align}
	we obtain three terms with different powers in $\beta$.
	
	In case of a conservative force, $\bbf = -\nabla_{\btheta} V$ this yields
	\begin{align}
		R & =\int_{0}^{T}\,\beta\,\langle||\nabla_{\btheta}V(\btheta)||^{2}\rangle-2\,\langle\Delta_{\btheta}V\rangle+\beta^{-1}\,\langle||\nabla_{\btheta}\ln p||^{2}\rangle\,dt,\label{eq:R_Vu_conservative}
	\end{align}
	where $\Delta_{\btheta}$ is the Laplace operator.

	\section{Derivation of NTK-related results\label{app:NTK}}
	
	Consider the gradient flow dynamics of the $i$-th parameter
	\begin{align}
		\frac{d\theta_i}{dt} &= - \sum_{\mu} \frac{\partial f_{\mu}}{\partial \theta_i} (f_{\mu} - y_{\mu})
	\end{align}
	where $f_{\mu}=f(\bm{x}_{\mu})$ uses NTK parameterization (i.e. weights of order $1$ and an explicit $1/\sqrt{width}$ factor accompanying pre-activations), and $y_\mu$ are the $\mu$-th target for $\mu \in [1,n]$. The learning rate is taken to be one, $\eta = 1$.
	
	Using SVD we can write 
	\begin{align}
		{\partial_{\theta_i} f_{\mu}} &= \sum_{\lambda \in \text{Spec[NTK]}} \sqrt{\lambda} u_{\mu,\lambda} v_{\lambda,i}
	\end{align}
	where $\lambda$'s are the NTK spectrum (times the width) and the vectors $u_{\mu,\lambda}$ or $v_{\lambda,i}$ for two different $\lambda$'s are orthogonal. The NTK matrix evaluated at two data points, $\mu, \nu$, is given by $\Theta_{\text{NTK}}(\bm{x}_\mu,\bm{x}_\nu) = \sum^{P}_{i=1}{\partial_{\theta_i} f_{\mu}}{\partial_{\theta_i} f_{\nu}}$. Multiplying the gradient flow equation with $v_{\lambda,i}$ and summing of $i$ one has
	\begin{align}
		\partial_t \theta_{\lambda} &= \sum_i v_{\lambda,i} \frac{d\theta_i}{dt} = - \sum_{i \mu} \sum_{\lambda' \in \text{Spec[NTK]}} \sqrt{\lambda'} (f_{\mu} - y_{\mu}) u_{\mu,\lambda'}  v_{\lambda,i} v_{\lambda',i} 
		=-\sqrt{\lambda} \Delta_{\lambda}(t)
	\end{align}
	where $f_{\lambda} = \sum_{\mu} f_{\mu} u_{\mu,\lambda}$ and, similarly with $y_{\lambda}$ and $\Delta_{\lambda}(t) \equiv f_{\lambda}(t) - y_{\lambda}$. The statement that the NTK does not change with training at infinite width, implies here that the SVD vectors and eigenvalue remain fixed. Furthermore, the original NTK derivation showed that 
	\begin{align}
		\Delta_{\lambda}(t) &= e^{- \lambda t} \Delta_{\lambda}(0)
	\end{align}
	plugging this into the last equation we obtain 
	\begin{align}
		\theta_{\lambda}(t) &= \theta_{\lambda}(0) + \frac{1}{\sqrt{\lambda}} \left[e^{-\lambda t}-1\right]\Delta_{\lambda}(0)
	\end{align}
	
	The Wasserstein-2 distance between a fixed initial state and the state at $t$ simplifies here to the $L_2$ distance, yielding 
	\begin{align}
		\sum_{\lambda} (\theta_{\lambda}(\infty)-\theta_{\lambda}(0))^2 &= \sum_{\lambda} \lambda^{-1} \left[e^{-\lambda t} -1\right]^2 \Delta_{\lambda}(0)^2
	\end{align}
	Next we note that $\beta^{-1} R$ simplifies here to the decrease in train loss indeed 
	\begin{align}
		\beta^{-1} R &= \eta^2 \int^{T}_0 dt (\nabla V)^2 + O(\beta^{-1}) = - \eta \int_{\btheta(0)}^{\btheta(T)} d {\btheta} (\nabla L) + O(\beta^{-1}) \\ \nonumber 
		&= - \eta [L(\btheta(T))-L(\btheta(0))] + O(\beta^{-1}).
	\end{align}
	
	How optimal are the NTK dynamics? To quantify this, we study the time bound over the actual training time, where the time bound is computed w.r.t. $\theta_{\lambda}(t)$. The advantage of such a quantity is that it is independent of the arbitrary learning rate (recall we already neglected discretization effects) and hence we can take it to $1$. Collecting the above results, this ratio is given by 
	\begin{align}
		\frac{T_{\text{SL}}(t)}{t} &= \frac{1}{t} \frac{\sum_{\lambda} \lambda^{-1} \left[e^{-\lambda t} -1\right]^2 \Delta_{\lambda}(0)^2
		}{\sum_{\lambda}\frac{\Delta_{\lambda}(0)^2}{2}-\sum_{\lambda}\frac{\Delta_{\lambda}(t)^2}{2}} = \frac{2}{t} \frac{\sum_{\lambda} \lambda^{-1} \left[1-e^{-\lambda t} \right]^2 \Delta_{\lambda}(0)^2
		}{\sum_{\lambda}\Delta_{\lambda}(0)^2\left[1-e^{-2\lambda t}\right]}
	\end{align}
	For a generic NTK kernel ($\Theta_{\text{NTK}}$) and any finite amount of data the ratio $\frac{T_{\text{SL}}}{t}$ decays as $\frac{2 \Delta(0)^T \Theta_{\text{NTK}}^{-1} \Delta(0)}{t |\Delta(0)|^2}$ for large enough $t$. This decay as $1/t$ signifies the fact that from some point onward, only exponentially weak (and hence negligible) learning is taking place. 
	
	Less general and more interesting results could be obtained by making some scaling assumptions on $\lambda$ and $\Delta_{\lambda}(0)$. Specifically, we assume $\lambda_{k} = \Lambda k^{-\alpha}$ and $\Delta^2_{\lambda_k}(0) = \Delta^2 k^{-\delta}$ where $k \in k_{\star}, \ldots, n$ for some $k_{\star}\ge 1$, and $\Delta,\Lambda,\alpha, \delta >0$. We further choose $t \approx \lambda_{n}^{-1}$
	such that many modes are learned, but some are still left to be learned. 
	
	Consider first the Wasserstein-2 term, 
	\begin{align}
		\sum_{k} \frac{\Delta^2_{\lambda_k}(0) [1-e^{-\lambda_k t}]^2}{\lambda_k} \approx \frac{\Delta^2}{\Lambda}\int_{k_\star}^{n} dk [1-e^{-\Lambda k^{-\alpha} t}]^2 k^{-\delta+\alpha},   
	\end{align}
	where our replacement of a summation by an integral is justified for high values of $k$ with an additional finite sum correction that is negligible in the limit of large $n$. As we will show, the contribution from high $k$ diverges with, $T$, and hence these dominate over the low $k$ part of the sum. Next making the substitution $x = \Lambda k^{-\alpha} t$ (or $k=[(\Lambda t)/x]^{\alpha^{-1}}$) we find 
	\begin{align}
		&\frac{\Delta^2}{\Lambda}\int^{\Lambda t}_{\Lambda t n^{-\alpha}} dx k^{\alpha+1} (\Lambda t)^{-1} [1-e^{-x}]^2 k^{-\delta+\alpha} 
		\\&= \frac{\Delta^2}{\Lambda}\int^{\Lambda t}_{\Lambda t n^{-\alpha}} dx [(\Lambda t)/x]^{\alpha^{-1}+2-\delta/\alpha} (\Lambda t)^{-1} [1-e^{-x}]^2 \\ \nonumber 
		&= \Delta^2 \Lambda^{\alpha^{-1}-\delta/\alpha} t^{\alpha^{-1}+1-\delta/\alpha}\int^{\Lambda t}_{\Lambda t n^{-\alpha}} dx x^{-\alpha^{-1}-2+\delta/\alpha} [1-e^{-x}]^2
	\end{align}
	Noting that $[1-e^{-x}]^2$ scales as $x^2$ at low $x$ the integral is non-divergent around its lower limit for $\alpha^{-1}(1-\delta)<1$, hence taking this lower limit to zero does not change the overall asymptotic. Furthermore, for, $\alpha^{-1}(1-\delta)+1>0$ the integral is convergent around the upper limit (which is in fact the lower limit of the original $dk$ integration). Hence, as far as the large $t$ asymptotic is concerned, we find 
	\begin{align}
		\mathcal{W}_2 &\rightarrow \left[\Delta^2 \Lambda^{\alpha^{-1}-\delta/\alpha} \int^{\infty}_{0} dx x^{-\alpha^{-1}-2+\delta/\alpha} [1-e^{-x}]^2 \right]t^{\alpha^{-1}+1-\delta/\alpha}
	\end{align}
	
	Next, we apply a similar line of reasoning to $\beta^{-1}R$:
	\begin{align}
		&\sum_{\lambda} \Delta_{\lambda}(0)^2 [1-e^{-2\lambda t}] = \sum_k \Delta^2 k^{-\delta} [1-e^{-2 \Lambda k^{-\alpha}t}] \approx \int_1^n dk \Delta^2 k^{-\delta} [1-e^{-2 \Lambda k^{-\alpha}t}]
	\end{align}
	using the same substitution of variables we have 
	\begin{align}
		\int^{\Lambda t}_{\Lambda t n^{-\alpha}} dx k^{\alpha+1-\delta} (\Lambda t)^{-1} [1-e^{-2 x}] &= \int^{\Lambda t}_{\Lambda t n^{-\alpha}} dx [(\Lambda t)/x]^{\alpha^{-1}+1-\delta/\alpha} (\Lambda t)^{-1} [1-e^{-2 x}] \\ \nonumber 
		&= (\Lambda t)^{\alpha^{-1}(1-\delta)} \int^{\Lambda t}_{\Lambda t n^{-\alpha}} dx [1/x]^{\alpha^{-1}+1-\delta/\alpha} [1-e^{-2 x}] 
	\end{align}
	Similarly to the Wasserstein-2 distance, for $\alpha^{-1}(1-\delta)<1$ the lower integration boundary is convergent. For $\alpha^{-1}(1-\delta)>0$ the top integration boundary is also convergent, leaving us with an $t^{\alpha^{-1}(1-\delta)}$ asymptotics. On the other hand, for $\alpha^{-1}(1-\delta)<0$ it is divergent and therefore leading to an additional $t^{-\alpha^{-1}(1-\delta)}$. Recalling that $\alpha > 0$ overall we find 
	\begin{align}
		\beta^{-1}R &\rightarrow \left[ \Lambda^{\alpha^{-1}(1-\delta)} \int_{0}^{\infty} dx [1/x]^{\alpha^{-1}+1-\delta/\alpha} [1-e^{-2 x}]  \right] t^{\alpha^{-1}(1-\delta)} \,\,\,\,\,\ (1-\delta) > 0 \\ \nonumber
		\beta^{-1}R &\rightarrow \left[ \Lambda^{\alpha^{-1}(1-\delta)} ] [\alpha^{-1}(\delta-1)]^{-1}  \right] t^{0} \quad \quad \quad \quad \quad \quad \quad \quad \quad \quad \quad \,\,\,\,\, (1-\delta) < 0
	\end{align}
	Collecting these results, one arrives at those of the main text.

	\subsection{Geometric length of NTK trajectory}
	Next, we address the geometry of the curve in the weights' space generated by the training procedure. 
	In general the length of a path $\vec{\gamma}$ parameterized by $\tau$ in Euclidean space is 
	\begin{align}
		l_{\gamma} &= \int_0^{\tau_{max}} d\tau \sqrt{ (\partial_{\tau} \vec{\gamma})^2 } 
	\end{align}
	in our NTK context $\tau = t$ and 
	\begin{align}
		\vec{\gamma}(t) &= \left(\frac{1}{\sqrt{\lambda_1}} \left[e^{-\lambda_1 t}-1\right]\Delta_{\lambda_1}(0),\frac{1}{\sqrt{\lambda_2}} \left[e^{-\lambda_2 t}-1\right]\Delta_{\lambda_2}(0),...\right)
	\end{align}
	where we recall that $\Delta_{\lambda}(0)=f_{\lambda}(0)-g_{\lambda}$ the latter being, respectively, network output and target projected on the $\lambda$ SVD eigenvector. The NTK trajectory length is thus  
	\begin{align}
		l_{\gamma} &= \int_0^{t} dt \sqrt{ \sum_k \lambda_k  e^{-2\lambda_k t}\Delta^2_{\lambda_k} }.
	\end{align}
	
	To see some explicit dependence on the spectrum, namely that it is power law $\lambda_k = k^{-\alpha}$ and that, $\Delta^2_{\lambda_k}=\Delta^2 k^{-\delta}$ with $\alpha> 0$, $,\delta\geq 0$ and hence independent of $k$. Following this, we approximate 
	\begin{align}
		l_{\gamma} &\approx |\Delta|\int_0^{t} dt \sqrt{ \int_1^{d} dk k^{-\alpha-\delta}  e^{-2 k^{-\alpha} t} } \\&= |\Delta|\int_0^{t} dt \sqrt{ \int_{2t d^{-\alpha}}^{2t} dx (2t \alpha)^{-1} (x/2t)^{-1/\alpha} (x/2t)^{\delta/\alpha} e^{-x} } \\ \nonumber 
		&= |\Delta|\int_0^{t} dt (2t \alpha)^{-1/2}(2t)^{1/(2\alpha)-\delta/(2\alpha)} \sqrt{ \int_{2t d^{-\alpha}}^{2t} dx  x^{-1/\alpha+\delta/\alpha}  e^{-x} },
	\end{align}
	where we used the change of variables $x = 2 k^{-\alpha} t$. At least for $\alpha > 1$, the $dx$ integration is non-singular at small $x$ hence for $t \ll 1/\lambda_{n}$, such that the lowest eigenmodes are non-learnable, we can replace the lower integration boundary by zero. Following this we obtain a lower incomplete gamma function 
	\begin{align}
		&|\Delta|\int_0^{t} dt (2t \alpha)^{-1/2}t^{1/(2\alpha)-\delta/(2\alpha)} \sqrt{ \int_{0}^{2t} dx  x^{-1/\alpha+\delta/\alpha}  e^{-x} } \\ \nonumber 
		&\equiv |\Delta|\int_0^{t} dt (2t \alpha)^{-1/2} t^{1/(2\alpha)-\delta/(2\alpha)} \sqrt{ \gamma(1-\alpha^{-1}+\delta/\alpha,2t) }
	\end{align}
	Notably at large $t$ (and correspondingly large $d$), the above integral is dominated by a $t^{(1+\alpha^{-1}-\delta/\alpha)/2}$ divergence as $\gamma(1-\alpha^{-1}+\delta/\alpha,2t) \rightarrow \Gamma(1-\alpha^{-1}+\delta/\alpha)$. No other factors, outside $O(1)$ factor, multiply this divergence. This divergence reflects the fact that the path gets longer as more and more modes are being learned. Examining potential divergences around $t=0$ (this time for the special case of $\delta=0$) one can expand around $t=0$ yielding 
	\begin{align}
		&|\Delta|\int_0^{t} dt (2t \alpha)^{-1/2}(2t)^{1/(2\alpha)} \sqrt{ \gamma(1-\alpha^{-1},2t) } \\ \nonumber 
		&=|\Delta|\int_0^{t} dt (2t \alpha)^{-1/2} (2t)^{1/(2\alpha)}\sqrt{ (2t)^{1-\alpha^{-1}} \Gamma(1-\alpha^{-1}) e^{-2t}\sum_{j=0}^{\infty} \frac{(2t)^j}{\Gamma(1-\alpha^{-1}+j+1)}} \\ \nonumber 
		&= |\Delta|\sqrt{\Gamma(1-\alpha^{-1})/\alpha} \int_0^{t} dt (2t)^{-1/(2\alpha)} (2t)^{1/(2\alpha)}e^{-t}\sqrt{\sum_{j=0}^{\infty} \frac{(2t)^{j}}{\Gamma(1-\alpha^{-1}+j+1)}} 
	\end{align}
	hence for $\alpha>0$ we see no low $t$ divergence. 
	
	These two results, especially the long $t$ divergence, should be compared with the $L_2$ distance of a straight-line trajectory at time $t$ given by 
	\begin{align}
		l^2_{\text{geo}} &= \sum_{\lambda} \frac{\Delta_{\lambda}(0)^2[1-e^{-\lambda t}]^2}{\lambda} \approx \sum_{k=0}^{t^{\alpha^{-1}}} \frac{\Delta_{\lambda_k}(0)^2}{k^{-\alpha}}
	\end{align}
	where we made a heuristic approximation and sharply separated learnable and unlearnable modes as those with $t \lambda > 1$ and $t \lambda < 1$ (specifically we took $[1-e^{-\lambda t}]^2$ to be $1$ for the former and zero for the latter).
	
	Next, making the same assumptions as those carried for the NTK trajectory, we find 
	\begin{align}
		l_{\text{geo}} &= |\Delta| \sqrt{\int_1^{t^{\alpha^{-1}}} dk k^{\alpha-\delta}} = |\Delta|\sqrt{(\alpha-\delta)^{-1} [t^{(\alpha-\delta+1)/\alpha}-1]}  
	\end{align}
	thus we find a divergence going as $t^{(\alpha^{-1}+1-\delta/\alpha)/2}$. Comparing both asymptotic we find 
	\begin{align}
		l_{\text{geo}}(t) &\propto t^{(\alpha^{-1}+1-\delta/\alpha)/2} \\ \nonumber 
		l_{\gamma}(t) &\propto  t^{(\alpha^{-1}+1-\delta/\alpha)/2}
	\end{align}
	Interestingly, we find the same asymptotic for the lengths, independent of $\alpha$ and $\delta$ (for  $\alpha>0,\delta>0$ and $(\alpha^{-1}+1-\delta/\alpha)>0$). 
	
	\section{Linear regression in high dimension\label{sec:linear_regression_hd}}
	
	Consider the problem of linear regression with scalar output, given a dataset $\mathcal{D}_n=\{\bm{x}_\mu,y_\mu\}_{\mu=1}^n = \{X,\bm{y}\}$ where $X\in \mathbb{R}^{d\times n}$, and $\bm{y}\in \mathbb{R}^{n}$. Our estimator for the output is a plugin estimator (student model)
	$\hat{{y}}(\boldsymbol{x};\boldsymbol{\theta})=\boldsymbol{\theta}^{\T}\boldsymbol{x}$. We aim to minimize the loss function  $\hat{\mathcal{L}}(\boldsymbol{\theta})=\frac{1}{2}\sum_{\mu}(y_{\mu}-\hat{{y}}(\boldsymbol{x}_\mu;\boldsymbol{\theta}))^{2}$, and find the optimal estimator for $\btheta$ via Langevin algorithm with learning rate $\eta$ 
	\begin{equation}
		d \btheta(t)   = -\eta \left(\sum_\mu(y_\mu -\btheta(t)^T\boldsymbol{x}_\mu)\boldsymbol{x}_\mu+\frac{\lambda d}{\beta} \btheta(t)\right)dt + \sqrt{2\eta \beta^{-1}} d\cB(t).
	\end{equation}
	The equilibrium distribution of this process matches the Bayesian posterior distribution which is independent of the learning rate.  To be more concrete, we evaluate the bound given the following
	noiseless target model \textbf{$\boldsymbol{y}_\mu=
		\boldsymbol{\theta}_{\star}^\T\bm{x}_\mu$} with
	$\boldsymbol{\theta}_{\star}\sim\mathcal{N}(0,\alpha/d I_{d}),$ and
	$\boldsymbol{x}_{\mu}$ are i.i.d. vectors with i.i.d. entries.

	In order to calculate the speed limit, we need to evaluate the Wasserstein-2 distance and the entropy production. Due to the linearity of this model, and the Gaussian assumption,
	all these quantities can be calculated exactly. In particular, both initial and final distributions are Gaussian, i.e., $\btheta_0 \sim p_{0}=\mathcal{N}(0,(\lambda d)^{-1}  I_{d})$,
	and $\boldsymbol{\theta}_T\sim p_{T}=\mathcal{N}(\boldsymbol{\mu}_{T},(\beta)^{-1}\Sigma_{T})$,
	where $\Sigma_{T}=\left(XX^{\T}+c_n I_{d}\right)^{-1}$, where $c_{n}= c d$, $c = \lambda /\beta$
	and $\boldsymbol{\mu}_{T}=\Sigma_{T}X\boldsymbol{y}$. 
	
	In the following, we take the leraning rate $\eta = 1/n$. Note that, the Wasserstein-2 is invariant to changes in the learning rate, whereas the $\beta^{-1}R$ will be affected by it. We start by calculating the partition functions, at initialization,  $\mathcal{Z}_0 = (2\pi/(\lambda d))^{d/2},$ and at the end of the training,  
	\begin{multline}
		\mathcal{Z}_{T}(\mathcal{D}_{n})
		=\int e^{-\frac{\lambda d}{2}\left\Vert \boldsymbol{\theta}\right\Vert ^{2}-\frac{\beta}{2}(\bm{y}-\boldsymbol{\theta}^{\T}X)^{\T}(\bm{y}-\boldsymbol{\theta}^{\T}X)}d\boldsymbol{\theta}\\
		=\left(|\Sigma_T|{\left(2\pi/\beta\right)^{d}}\right)^{1/2}e^{-\frac{\beta}{2}\|\boldsymbol{y}\|^{2}+\frac{1}{2}\beta\boldsymbol{y}^{\T}X^{\T}\Sigma_{T}X\boldsymbol{y}}    
	\end{multline}
	The entropy production (\eqref{eq:R_learning}) is then, 
	\begin{gather}
		(n\beta)^{-1}R=(n\beta)^{-1}\log\mathcal{Z}_{T}(\mathcal{D}_{n})-(n\beta)^{-1}\log\mathcal{Z}_{0}+\frac{1}{n} \langle\mathcal{L}(\btheta(0))\rangle\\\nonumber
		=\frac{\gamma_n}{2\beta }\log(c d)+\frac{1}{2\beta n}\log|\Sigma_{T}|
		+\frac{1}{2n}\|\Sigma_T^{-1/2}\mu_T\|^2
		+\frac{1}{2\lambda d n }\mathrm{Tr}\left(XX^{\T}\right)
		\\ \nonumber
		=\frac{\gamma_n}{2\beta}\log(c)-\frac{1}{2n\beta }\log|c I_{d}+\frac{1}{d}XX^{\T}|\\+\frac{1}{2d^2 n}\mathrm{Tr}\left(\left(c I_{d}+\frac{1}{d}XX^{\T}\right)^{-1}\left(XX^{\T}\right)^{2}\btheta_\star \btheta_\star^\T\right)+\frac{1}{2\lambda 
			d n}\mathrm{Tr}\left(XX^{\T}\right),
	\end{gather}
	where $\gamma_n = d/n$. Since both distributions at initialization and at the end of training are Gaussian, the Wasserstein distance can be calculated exactly, 
	\begin{multline}
		W^{2}(p_{0,}p_{T})
		=\left\Vert \mu_{0}-\mu_{T}\right\Vert ^{2}+\mathrm{Tr}\left(\Sigma_{0}+\beta^{-1}\Sigma_{T}-2\beta^{-1/2}\left(\Sigma_{T}^{1/2}\Sigma_{0}\Sigma_{T}^{1/2}\right)^{1/2}\right)\\ \nonumber
		=\left\Vert \mu_T \right\Vert ^{2}+\lambda^{-1}+\beta^{-1}\mathrm{Tr}\left(\Sigma_T\right)-2(\beta\lambda )^{-1/2}d^{-1/2}\mathrm{Tr}\left(\Sigma_T^{1/2}\right)
		\\ \nonumber
		= \frac{1}{d^2}\mathrm{Tr}\left(\left(c I_{d}+\frac{1}{d}XX^{\T}\right)^{-2}\left(XX^{\T}\right)^{2}\btheta_\star \btheta_\star^\T\right)+\lambda^{-1}+\frac{1}{\beta d}\mathrm{Tr}\left(\left(c I_{d}+\frac{1}{d}XX^{\T}\right)^{-1}\right)\\-2(\beta\lambda)^{-1/2}d^{-1}\mathrm{Tr}\left(\left(c  I_{d}+\frac{1}{d}XX^{\T}\right)^{-1/2}\right).
	\end{multline}
	where $\Sigma_0 = 1/(\lambda d)I_d$ The speed limit bound, \eqref{eq:speed_limit},  is then,
	\begin{equation}
		\begin{gathered}
			T(\mathcal{D}_{n})\geq\frac{W^{2}(p_{0,}p_{T})}{\beta^{-1}R}
			\equiv T_\mathrm{SL}
		\end{gathered}    
	\end{equation}
	We note that the analysis here can be generalized to other data distributions (see section \ref{app:NTK}). In the regime, where $\gamma_n = d/n\to\gamma\in(0,\infty)$,
	and $d,n\to\infty$, the results simplify. 
	Taking expectation over $\btheta_\star$ and using the concentration of quadratic forms, the speed limit bound is, then, 
	
	\begin{multline}\label{eq:Tsl}
		T_{\text{SL}} =
		\frac{\lambda^{-1}+\alpha\int\left(c\gamma+s\right)^{-2}s^{2}d\rho(s)+\beta^{-1}\int\left(c+s/\gamma\right)^{-1}d\rho(s)}{\frac{\gamma}{2\beta}\log(c)-\frac{\gamma}{2\beta}\int\log|c+s/\gamma|d\rho(s)+\frac{1}{2\lambda}\int sd\rho(s)}
		\\
		-\frac{2(\beta\lambda)^{-1/2}\int\left(c+s/\gamma\right)^{-1/2}d\rho(s)}{\frac{\gamma}{2\beta}\log(c)-\frac{\gamma}{2\beta}\int\log|c+s/\gamma|d\rho(s)+\frac{1}{2\lambda}\int sd\rho(s)}
		+o(1)
	\end{multline} 
	where $\rho$ here is the limiting measure of the eigenvalues of $\frac{1}{n}XX^{T}$ for i.i.d entries and samples, known as the Marchenko--Pastur
	distribution. 
	It takes the following form $$\rho(x)=\begin{cases} (1-\frac{1}{\gamma}) \delta({x}) + \nu(x),& \text{if } \gamma >1\\
		\nu(x),& \text{if } 0\leq \gamma \leq 1,
	\end{cases}
	$$ with, 
	$\nu(x) = {\frac{1}{2\pi}}{\frac{\sqrt{(\gamma_{+}-x)(x-\gamma_{-})}}{\gamma x}}\,\mathbf{1} _{x\in[\gamma_{-},\gamma_{+}]}
	$, such that, $\gamma_{\pm} = (1\pm \sqrt{\gamma})^2$ where $\delta(x)$ is the Dirac delta function.
	
	Interestingly, taking the limit of ${\beta}\rightarrow\infty$ in \eqref{eq:Tsl} (note that $c\to0$, because $c=\lambda/{\beta}$). The speed limit is then
	\begin{equation} \label{eq:T_beta_infty}T_{\mathrm{SL}}({\beta} \to \infty )\to \frac{\lambda^{-1}+\alpha}{\frac{1}{2\lambda}\int sd\rho(s)}. 
	\end{equation}
	Note that, taking now the limit of $d\to \infty $ ($\gamma \to \infty$) corresponds to $d\gg n$, we get that $$\lim_{d \to \infty}\lim_{\beta \to \infty}T_{\text{SL}}= 2({1+\alpha\lambda}).$$ In this regime, the weight decay term generates additional noise due to over-parametrization.  
	
	On the other hand, if we take ${\beta}\to 0$ we get
	\begin{equation}\label{eq:T_beta_0}
		T_{\mathrm{SL}}({\beta} \to 0) \to 0.
	\end{equation} 
	In this regime, the system is driven by noise, and there is essentially no learning. In the over-parametrized regime in which $d\to \infty$, and $d\gg n$ ($\gamma \to \infty $) we have that
	\begin{equation}
		T_{\mathrm{SL}}(d \to \infty )  \to 0
	\end{equation} 
	This shows that when the system is extremely over parametrized the distribution is barely moving from its initial condition. We note that as shown above that will not be the case in zero noise. I.e. the limit of $d\to \infty$ does not commute with the limit of $\beta\to \infty$. 
	
	Last, as $n\to\infty$, ($\gamma \to 0$) corresponds to $n\gg d$, the bound reaches the following finite value: 
	\begin{equation} 
		T_{\mathrm{SL}}(n\to \infty) \to 2\lambda \alpha.
	\end{equation} 
	This limit is in essence where we learn the population error i.e. the expectation of the loss function over the true dataset distribution. Remarkably, it is independent of the amount of noise $\beta$.
\end{document}